\documentclass{isprs} % isprs class modified 23-04-2019 (Dennis Wittich)
\usepackage{subfigure}
\usepackage{setspace}
\usepackage{geometry} % added 27-02-2014 Markus Englich
\usepackage{epstopdf}
\usepackage[labelsep=period]{caption}  % added 14-04-2016 Markus Englich - Recommendation by Sebastian Brocks
\usepackage[british]{babel} 
\usepackage[hang]{footmisc}
\usepackage{amsmath} % Jan Rudolph 20.10.2025
\usepackage{amssymb} % Jan Rudolph 23.10.2025
\usepackage{booktabs} % Jan Rudolph 23.10.2025
\usepackage{comment} % Jan Rudolph 23.10.2025
\usepackage{hyperref}
\usepackage[authoryear]{natbib}
\newcommand{\Unit}[1]{\thinspace \mathrm{#1}}
\begin{document}
\title{A Novel Camera-to-Robot Calibration Method for Vision-Based Floor Measurements}
\date{}
\author{Jan Andre Rudolph, Dennis Haitz, Markus Ulrich}
\address{Machine Vision Metrology Lab, \\Institute of Photogrammetry and Remote Sensing, Karlsruhe Institute of Technology, Germany\\\{jan.rudolph, dennis.haitz, markus.ulrich\}@kit.edu}

\abstract{
A novel hand–eye calibration method for ground-observing mobile robots is proposed. While cameras on mobile robots are common, they are rarely used for ground-observing measurement tasks. Laser trackers are increasingly used in robotics for precise localization. A referencing plate is designed to combine the two measurement modalities of laser tracker 3D metrology and camera-based 2D imaging. It incorporates reflector nests for pose acquisition using a laser tracker and a camera calibration target that is observed by the robot-mounted camera. The procedure comprises estimating the plate pose, the plate–-camera pose, and the robot pose, followed by computing the robot–-camera transformation. Experiments indicate sub-millimeter repeatability.
}
\maketitle

\section{Introduction}\label{sec:introduction}
\sloppy
\paragraph{} Robot systems have been utilized in industrial facilities for decades to automize production.  
However, such automation strategies were targeted at the transport of products within the manufacturing process. 
For specialized manufacturing tasks, in the 1960s and onwards~\citep{Zamalloa2017,Gasparetto2019} robot arms began to emerge as a new automation strategy. 
Other types of robot systems with fewer degrees of freedom are used for tasks like pick-and-place, sorting or machine loading~\citep{Gasparetto2019}.

\paragraph{} In recent years, a different type of robot system has been introduced in industrial manufacturing in the form of mobile platforms. In manufacturing environments and especially logistics, ground-based systems are increasingly used for transportation tasks~\citep{PerezGrau2021JMS,Tehrani2022IJIRA}. 
Those ground-based mobile platforms can be categorized into path-dependent, e.g. by following markers on the ground detected through optical sensing, and fully autonomous systems. 
Often, the positioning of such systems is implemented using inertial measurement units (IMU) or with optical sensors like cameras or LiDAR. 
Both optical sensors and IMUs can also be used within an integrated setup to compute the position, orientation, and environment map within a Simultaneous Localization and Mapping (SLAM) algorithm, or visual SLAM, if optical sensors are available.

\begin{figure}
    \centering
    \includegraphics[width=0.7\linewidth]{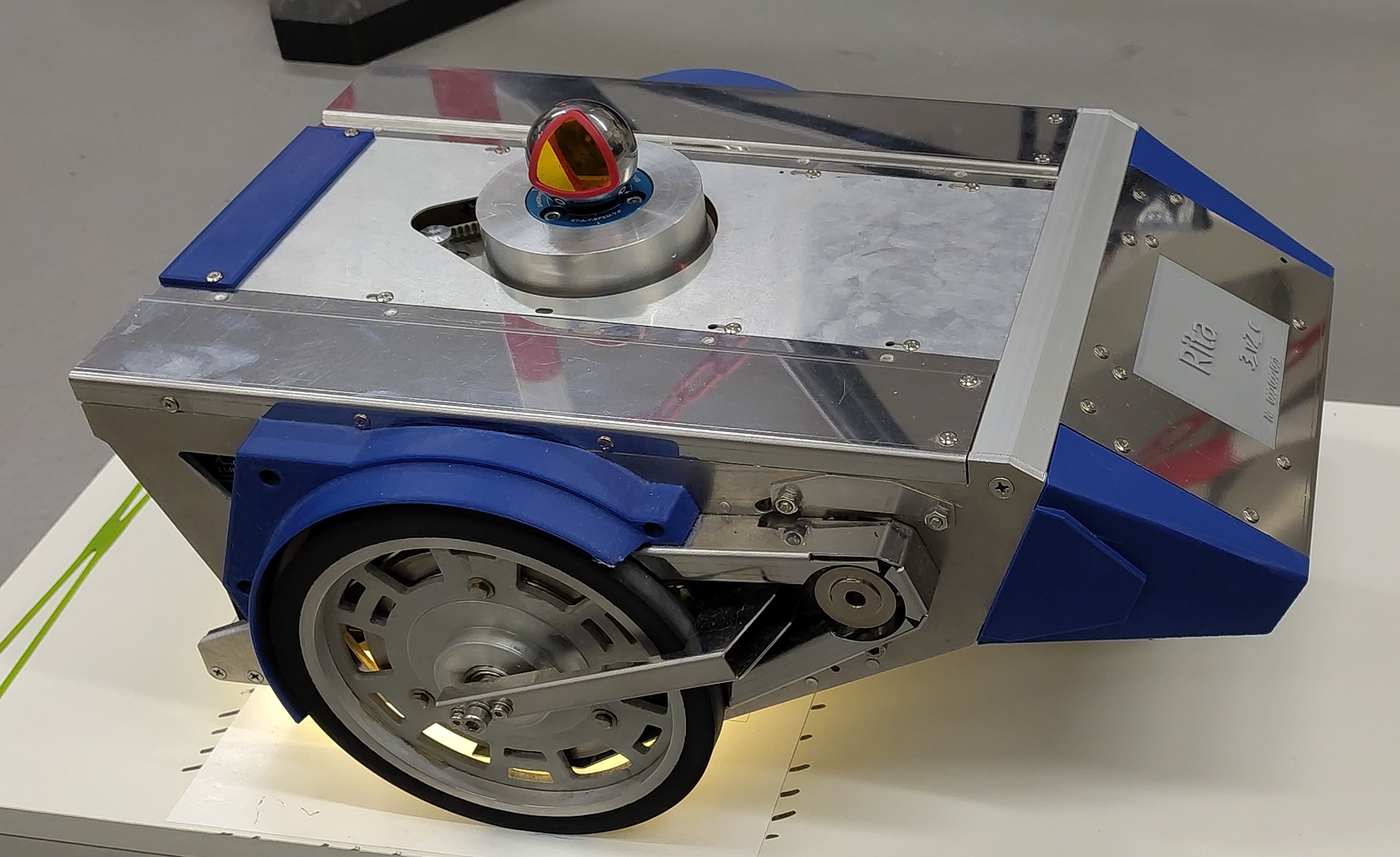}
    \caption{Mobile robot Rita (adapted from \cite{Naab+2023+109+117}), used in the proposed experiments. It features a differential drive with a front-mounted caster wheel. A spherically mounted retro-reflector is installed on top within a rotatable mount, maintaining the line of sight to a laser tracker. The nadir-mounted camera remains occluded in this image, with only its active illumination apparent.}
    \label{fig:rita}
\end{figure}

\paragraph{} While ready-to-use mobile platforms often include positioning systems and ground-facing cameras, these are usually used for path detection. 
There are tasks, however, which require high precision positioning of a mobile system w.r.t.\ ground-based markers. 
The mobile \textit{Robot with Integrated Tracker steering for different Applications} (RITA, Fig.~\ref{fig:rita}) has been developed for sub-millimeter precise positioning w.r.t.\ known marker positions on the ground of large production and logistics facilities~\citep{Naab2025RITA}. 
Marking and detecting those markings for high precision ground drillings is its core development objective. A stationary laser tracker as a polar measurement system is used to accurately determine the position of a platform-mounted reflector with a distance measurement precision of $10 \Unit{\mu m/m}$. 
\paragraph{} To detect the marking on the ground, a ground-facing, calibrated RGB camera with industrial specifications is built into the mobile platform. 
Acquired images of the ground below the platform are then utilized to determine the coordinates of a marking in image space.
The objective of this research is to transform these image coordinates from the camera coordinate system (CCS) to the robot coordinate system (RCS), which requires a relative orientation between both systems.
Over the course of development, we discovered that proven methods for camera-to-robot calibration were not applicable for our case.
Therefore, in this work, we present a specialized calibration procedure. 
We introduce a standard circular grid calibration plate, which is customized with three colored (red, green, and blue) circles and a reflector within each of the three circles. 
The mobile platform moves upon the ground laying plate to a first position and acquires an image of the plate, whereby the laser tracker measures the position of the platform by using the reflectors. 
Then, the mobile platform is moved to a second position and the laser tracker measures the new position of the platform, as well as the position of the three calibration-plate-mounted reflectors, which allows for the computation of the relative position of the colored circles in the CCS w.r.t.\ the platform-mounted reflector in the RCS.

In short, we present the following contributions:
\begin{itemize}
    \item A customized calibration plate that can be used for camera as well as laser tracker measurements in our system setup
    \item A calibration procedure to establish the relative orientation between the CCS and RCS using camera and laser tracker.
\end{itemize}

\section{Related Work}\label{sec:related-work}
This section provides an overview of mobile robot systems in the context of industrial environments, which includes production and logistics. 
\paragraph{Ground operating mobile robots}
Ground operating mobile robots can be categorized into two categories: Automated Guided Vehicles (AGVs) and Autonomous Mobile Robots (AMRs). \citet{KeithLa2024} identify the ability of decision making of AMRs as the main difference between both systems. \citet{Oyekanlu2020} on the other hand, use the term AMR to characterize various types of robot systems, e.g., an AGV can also be an AMR, if it meets certain autonomy criteria. They also note that both terms in literature are often used interchangeably. \citet{ZhaoChidambareswaran2023CASE} identify four key differences, which apply to AMRs: Independent movement ("Go Anywhere"), diversified applications, high adaptation capabilities ("Zero disruption to production"), and cost effectiveness.
To navigate through an environment, landmark-based approaches with robot-mounted cameras have been a long-studied approach, besides lidar and laser triangulation methods~\citep{HuGu2000,Shneier2015NIST}. Other than landmarks, robot position can be obtained from known and detected objects based on sensor data~\citep{BenAriMondada2018}. Another method for AGV routing is realized through ground-fixed magnetic tape, inductive wire~\citep{Shneier2015NIST,ZhaoChidambareswaran2023CASE} or laser projection~\citep{Tsuruta2019AutCon}. For both AMRs and AGVs, SLAM-based approaches are widely used, if no information like CAD models about the environment is provided~\citep{Harapanahalli2019,WangZhang2024SLAM}. If key objects in the environment are known, e.g.\ through ground truth images, a (visual) SLAM algorithm can be used for independent navigation, if an object detection algorithm is provided~\citep{Asadi2018AutCon}. The key difference is that AMRs navigate independently, while AGVs strictly follow predefined paths~\citep{Zhang2023AutCon,ZhaoChidambareswaran2023CASE}.

\paragraph{Optical robot calibration}
In this section, we only include calibration methods that relate camera coordination systems to robot coordination systems. 
Calibration of camera intrinsics~\citep{Zhang2000CameraCalib}  as well as kinematic calibration without optical sensors~\citep{Li2021Sinica} are excluded.

\paragraph{Hand--eye calibration} Hand--eye calibration describes the determination of the camera pose relative to the robot~\citep{Steger2018Book}. Initially, early hand--eye calibration methods from~\citet{LenzTsai1988CVPR} and~\citet{TsaiLenz1989} introduced the terminology of eye-on-hand configuration, whereby the camera is mounted on the last joint of a robot arm with a calibration body or plate in the robot workspace. Those and following methods~\citep{ChouKamel1988ICRA,ParkMartin1994} first determine rotation, then translation. Simultaneously determining rotation and translation was achieved by follow-up works~\citep{Daniilidis1996ICPR,Daniilidis1999IJRR,StroblHirzinger2006IROS}, mainly using dual quaternions to avoid non-linear optimization. \citet{Ulrich2024ISPRS} simultaneously calibrate robot kinematics, the hand--eye transformation, and the intrinsic camera parameters with a robot-mounted camera. \citet{UlrichHillemann2021ICRA,UlrichHillemann2024TRO} approach the challenge of uncertainty between the precision of the robot and the absolute accuracy from a geodetic point of view. These methods have in common that they require non-parallel screw axes between the calibration poses. In case of a planar mobile robot motion, all screw axes are orthogonal to the floor and therefore parallel to each other. A similar challenge arises in the hand-eye calibration of SCARA robots~\citep{ISO8373_1994,ISO9787_1999}, whose rotational joints are all parallel. In such cases, the hand--eye pose can be determined only up to an unknown translation along the direction of the rotation axes.  \citet{UlrichSteger2016PRIA} propose a solution for SCARA hand--eye calibration that also determines the missing translation component by moving the robot tool center point to a predefined height. This approach cannot be applied to the mobile robot, as it would require a vertical movement of the robot and physical access the center of the retro-reflector.

\paragraph{Laser trackers} Laser trackers are a common optical measurement system used for kinematic robot calibration~\citep{Abderrahim2006Intech,Chen2014TRO,Li2016TRO,Liu2018IndRobot}, especially for high-precision determination of the position of a robot end-effector. \citet{Li2021Sinica} lay out the limitations of such methods, whereby the main disadvantage is that the robot can create self-occlusion w.r.t.\ the stationary laser tracker. Because a laser tracker is used for kinematic calibration only,~\citet{Ulrich2024ISPRS} point out that a camera-based calibration of the kinematics can be a cost-effective and easy alternative to the calibration using a laser tracker, if a camera is already integrated into the system configuration.

\paragraph{Floor marking and detection}
To automatically mark stand positions for exhibitions and trade fairs,~\cite{Jensfelt2006JFR} describe an AGV that marks a predefined set of positions on the ground in conjunction with a CAD model of a large facility like an event hall. The average absolute positional accuracy is $28\Unit{mm}$ with a standard deviation of $18\Unit{mm}$. Utilized for indoor construction sites but closely related to the objective of the Rita system,~\citet{Tsuruta2019AutCon} introduced a mobile robot system for floor marking, that follows a laser-projected line on the ground. The robot itself contains cameras to detect the projected lines. Regarding accuracy, the authors state the average deviation (w.r.t.\ x- and y-direction) with $2.3\Unit{mm}$, and the standard deviation with $1.4\Unit{mm}$. \citet{Iqbal2023Buildings} present a system with an accuracy of 10 to $15\Unit{mm}$ for a line drawing task.
There exists a wide range of research for mobile robot positioning based on fiducial markers that are attached to the wall or ceiling~\citep{MondejarGuerra2018,RomeroRamirez2018IVC,Fiala2009TPAMI}, or especially ArUco markers~\citep{Zheng2018ROBIO,Filus2022BigData}. \citet{Gwak2021Sensors} propose two special QR-code-like markers, namely hierarchical and nested markers, for horizontally moving mobile robots (e.g., AGVs) as well as robots with the ability to move vertically (e.g., UAVs), respectively. The problem of robot positioning is closely related to camera pose estimation, which is a fundamental photogrammetric problem and typically solved by spatial resection and bundle adjustment while exploiting the epipolar constraint for stereo pose initialization within a multi-view setup. These methods are used in modern Structure-from-Motion (SfM)~\citep{Schoenberger2016CVPR} or visual SLAM~\citep{MurArtal2015TRO} algorithms, however, with the constraint of identifying matching keypoints in corresponding images or keyframes.
Works by~\citet{Nguyen2017MTAP} and~\citet{Zhang2021AutSinica} focus on the health-related task of visual-impairment, using a mobile robot as an assistive system with integrated path planning. Especially cameras and active sensors are utilized, while Odometry and SLAM are used for positioning and mapping. Line and lane detection are further tasks for robot systems with onboard optical sensors, especially for the task of autonomous driving~\citep{Gajjar2023ESA} and for AGVs in industry~\citep{Liu2020ICMA,Zhang2023EEBDA}.
The Mark.One robot as a commercial product is a robot system for floor mark detection, which achieves a positioning accuracy of $1\Unit{mm}$~\citep{Solid3D2025MarkOne} using a built-in ground-facing camera. Additionally, a stationary laser tracker is included in the system configuration.

\section{Methodology}\label{sec:methodology}
In this section, the hardware-induced conditions are laid out with the design of a customized calibration plate as one major contribution. Method concerning assumptions w.r.t.\ the hardware setup are introduced at first occurrence. 
\paragraph{Coordinate systems} All \emph{coordinate systems (CS)} are assumed to be Euclidean and orthonormal. Cartesian coordinates are used. A point ${}^{A}P_i$ with index $i$ w.r.t.\ CS $A$ is a 3-tuple with homogeneous coordinate representation 
\begin{align}
    {}^{A}\bar{P}_i = 
    \begin{pmatrix}  
        {}^{A}P_i \\ 1
    \end{pmatrix}.
\end{align} In the following, a homogeneous transformation from CS A to B, denoted by $^{B}H_A$, is expected to be a $4 \times 4$ homogeneous matrix representing a rigid transformation, i.e. ${}^{B}H_A \in SE(3)$. The analogous notation for rotation matrices is ${}^{B}R_{A} \in SO(3)$, while ${}^{B}\mathbf{v}$ is a 3-tuple vector $\mathbf{v}$ w.r.t. B using a small letter without right hand side index.

\subsection{System Setup} \label{sec:hardware_context} 
A multitude of hardware components provide internal canonical coordinate frames. In the following, a precise notation of the used coordinate frames will be emphasized.
\paragraph{Absolute System (ACS, abs)} \label{par:lasertracker} Laser trackers accurately measure $3D$ coordinates of a moving target in a high frequency. In mid-range applications below 5 meters, the Absolute Tracker AT901's accuracy is specified below $35\Unit{\mu m}$ \citep{leicageosystemsLeicaAbsoluteTracker}.
Because this accuracy is more than an order of magnitude higher than the expected accuracy of the results, laser tracker measurements $^{abs}P$ can be utilized as ground truth. Local laser tracker coordinates are transformed into a global coordinate frame, denoting the ACS.
\paragraph{Reflector (smr)} 
We use spherically mounted retro-reflectors (smr) defined by their center point $^{abs}P_{smr}$ and radius. These are provided as a dedicated laser tracker measurement target by the manufacturer and mounted into a reflector nest. While simple ring-shaped steel reflector nests do not provide ideal conditions for $\Unit{\mu m}$ precise measurements, they are sufficient for sub-$\Unit{mm}$ precise measurements and offer a favorable shape for automatic detection in images.
\paragraph{Robot (RCS, rob)} As a reference system, the differential-drive robot Rita \citep{Naab+2023+109+117, Naab2025RITA} is used. Though the left and right drive wheels as well as the caster wheel are moving parts, the respective wheel contact points with the ground are modeled constant in the robot frame:
\begin{align}
    {}^{rob}P_{w} &= \mathrm{const}, \\
    w &\in \left\{ w_{\text{left}},\, w_{\text{right}},\, w_{\text{passive}} \right\}.
\end{align} 
Furthermore, the robot possesses an smr which is always turned into the direction of the laser tracker. The movement of the smr in the robot chassis is expected to be in sub-mm range, so it can be modeled as the RCS origin:
\begin{align}
    {}^{abs}P_{rob} &= {}^{abs}P_{smr}, \label{eq:robGleichSmr}\\
    {}^{rob}P_{smr} &= \begin{pmatrix} 0&0&0 \end{pmatrix}^\top.
\end{align} The three wheel contact points ${}^{abs}P_{w}$ do also define an ideal support plane underneath the robot. Based on the support plane normal ${}^{abs}\mathbf{n}$ (upside), the heading direction ${}^{abs}\mathbf{v}$ which yields ${}^{abs}\mathbf{v}_\perp$ as the normalized orthogonal projection of ${}^{abs}\mathbf{v}$ to a normal of ${}^{abs}\mathbf{n}$ and ${}^{abs}\mathbf{c} = {}^{abs}\mathbf{n} \times {}^{abs}\mathbf{v}_\perp$ completing a right-handed orthonormal basis, the robot orientation
\begin{align}
    {}^{abs}R_{rob} = \begin{bmatrix} {}^{abs}\mathbf{v}_\perp & {}^{abs}\mathbf{c} & {}^{abs}\mathbf{n} \end{bmatrix} \in SO(3)
    \label{eq:rob_R_basisDecomposition}
\end{align} is defined in the ACS. Combining Eqs.\@~(\ref{eq:robGleichSmr}) and (\ref{eq:rob_R_basisDecomposition}) yields:
\begin{align}
    {}^{abs}H_{rob} =
    \begin{bmatrix}
    {}^{abs}R_{rob} & {}^{abs}P_{rob}\\
    0 & 1
    \end{bmatrix}.
    \label{eq:composite_rob_H}
\end{align}

\paragraph{Camera (CCS, cam)}\label{par:camera} A ground-facing camera system is rigidly attached inside the robot. Consisting of the 12.2 MP Phoenix PHX122S model \citep{lucidvisionlabsPhoenix} with a 3.5mm ruggedized Techspec lens \citep{edmundopticsTechspec} it observes a ground area spanning over a few $\Unit{cm^2}$ on the floor underneath the robot. At this point, the associated camera model is assumed to be calibrated in advance. A ruggedized lens has been selected that preserves this calibration over time and is not affected by shocks or vibrations. This leads to the assumption that the CCS is fixed in the RCS:
\begin{align}
    {}^{rob}H_{cam} = \mathrm{const}.
\end{align}

\begin{figure}[t]
  \centering
  \subfigure[In-house-designed wooden referencing plate demonstrator which is not sufficiently flat and can be deformed by the mobile robot weight. Green and yellow rubber band is applied against slipping.\label{fig:woodplate}]{
    \includegraphics[width=1\linewidth]{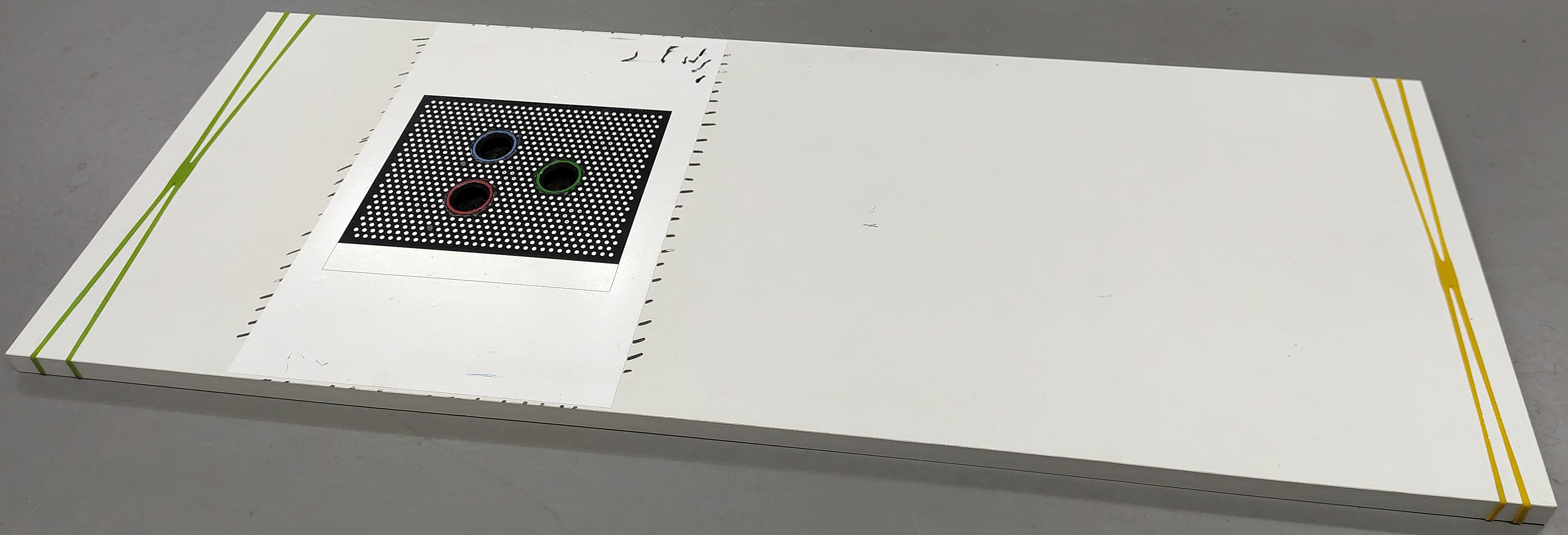}
  }
  \subfigure[Bespoke referencing plate which is made of $19mm$ thick glass providing a flat and stable surface.\label{fig:glassplate}]{
    \includegraphics[width=1\linewidth]{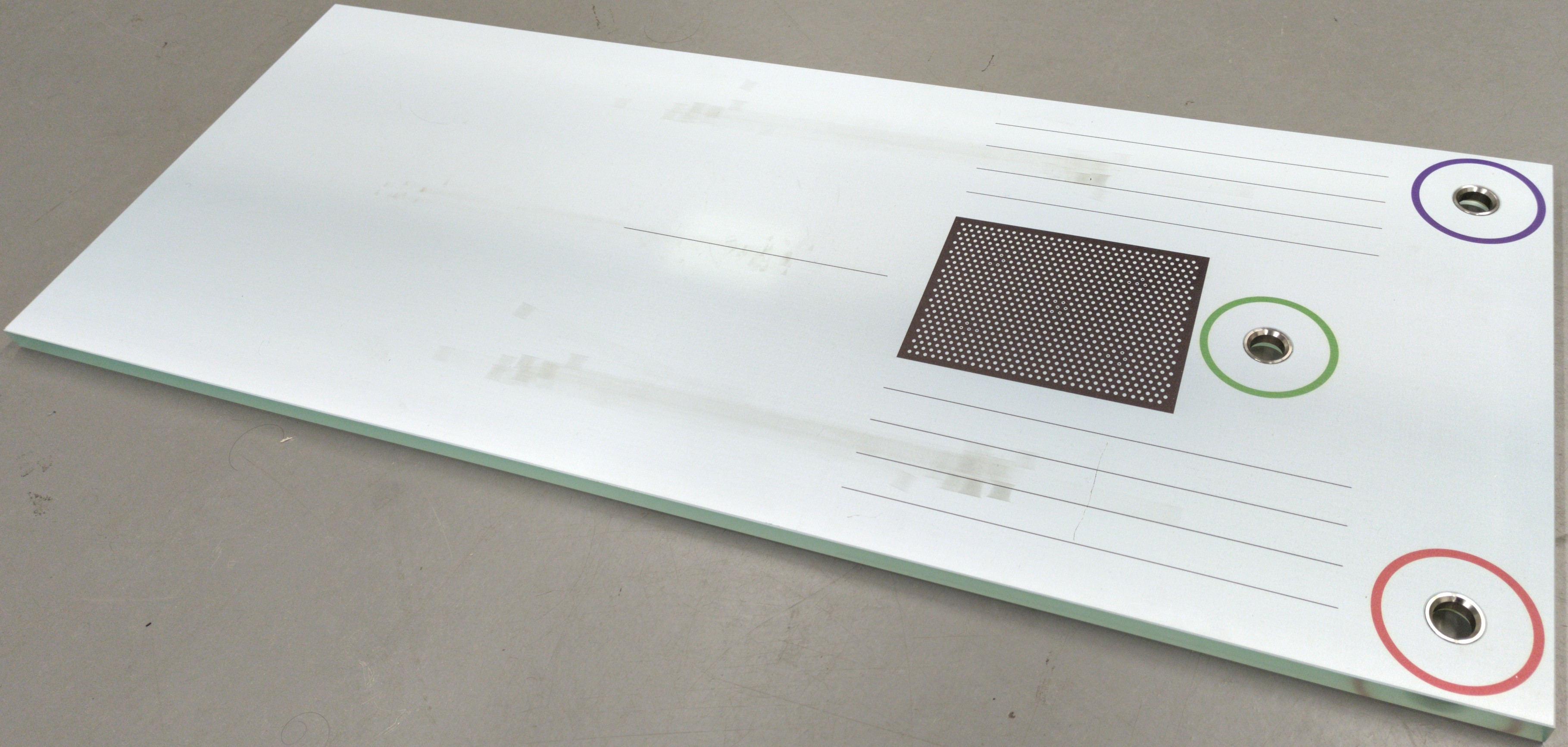}
  }
  \caption{Referencing plates used in experiments. While they are arbitrarily posed on the ground, a mobile robot Rita can be placed on them. The colored circles mark flush-mounted reflector nests. The dark texture represents the HALCON calibration target with circular marks. The rest of the white area provides space to move the robot for a few decimeters. }
  \label{fig:referencingplates}
\end{figure}

\paragraph{Referencing Plate (PCS, ref)} A referencing plate (Fig. \ref{fig:referencingplates}) was designed to combine the two measurement modalities of laser tracker 3D metrology and camera-based 2D imaging. On the one hand side, it provides reflector nests to obtain the pose in the ACS. On the other hand side, it integrates a camera calibration target on the surface, which can be measured by the camera when the robot moves over the plate. For this, we use a calibration target~\citep{MVTecHALCON2511-misc} which supports determining a unique pose by coded markers. The corresponding rigid transformation is denoted by ${}^{cam}H_{ref}$. The material ensures that the surface conforms to a plane within sub-millimeter tolerances and remains undeformed under robot load. Reflector nests ring surfaces are expected to be even or at least parallel to the surface. The surface must be on the xy-plane of the PCS, which is given by the integrated calibration target. This will also be the reference system for the referencing plate.

\newpage
\subsection{Referencing}
\begin{figure}
    \centering
    \includegraphics[width=1\linewidth]{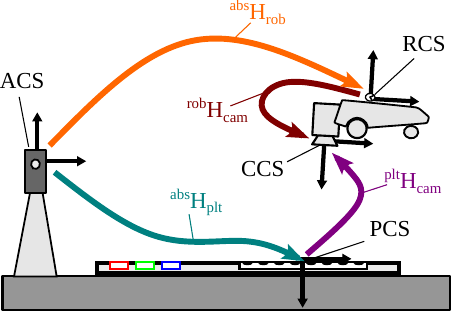}
    \caption{Retrieval of the orientation of the camera inside the robot frame. The plate-absolute-transformation ${}^{abs}H_{plt}$ and the plate-camera-transformation ${}^{plt}H_{cam}$, as well as the robot-absolute-transformation ${}^{abs}H_{rob}$ on the other side, are combined to robot-camera-transformation ${}^{rob}H_{cam}$ (figure adapted from \cite{UlrichHillemann2024TRO}).}
    \label{fig:idea}
\end{figure}

The core idea of our referencing method is to retrieve
\begin{itemize}
    \item the camera pose relative to the referencing plate,
    \item the plate pose in the abs system and
    \item the robot pose in the abs system,
\end{itemize}
when the robot is placed on the referencing plate such that the integrated calibration target is visible in the camera image. The transformations are visualized in Fig.\@~\ref{fig:idea}.
\begin{figure}
    \centering
    \includegraphics[width=1\linewidth]{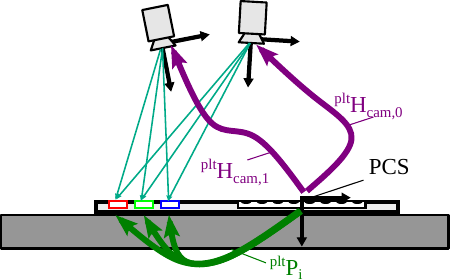}
    \caption{Stereo measurement of the reflector nest positions in the referencing plate frame. The plate-camera-transformations ${}^{plt}H_{cam,0}$ and ${}^{plt}H_{cam,1}$ are used to calculate the reflector nest positions ${}^{plt}P_{i}$ by triangulation (turquoise arrows) (figure adapted from \cite{UlrichHillemann2024TRO}).}
    \label{fig:stereomeasurement}
\end{figure}
\begin{figure}
    \centering
    \includegraphics[width=1\linewidth]{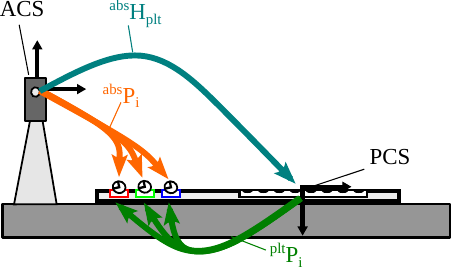}
    \caption{Retrieval of the absolute orientation of the referencing plate. Absolute laser tracker measurements ${}^{abs}P_{i}$ are matched to the reflector nest positions w.r.t. the referencing plate, yielding the absolute-plate-transformation ${}^{abs}H_{plt}$  (figure adapted from \cite{UlrichHillemann2024TRO}).}
    \label{fig:plate_pose}
\end{figure}

\paragraph{Scene (SCS, scn)} The scene describes a concept in the context of image rectification. Polynomial camera model parameters and ${}^{cam}H_{ref}$ are sufficient to project points from the image plane into the PCS-xy-plane. The SCS is chosen such that the projection of the entire image of the camera fits into the positive xy-quadrant and x- and y-axis are aligned to image row and column (z-axis pointing upwards now, contrarily to PCS). The calculation of the rectification map
\begin{align}
    \mathrm{map}^{scn\text{-}xy}_{image}: \mathbb{R}^2 \mapsto \mathbb{R}^2,
    \label{eq:rectification_map}
\end{align}
that maps image points directly into positive SCS-xy-plane yields ${}^{scn}H_{cam}$. In this contribution, image-based measurements are directly represented in the SCS. The conceptual advantage is the following: While the PCS is bound to the referencing plate, the SCS is bound to the CCS and therefore the RCS, even when the robot moves. It directly denotes the location of the rectified image on the floor, assuming planarity in between the wheel contact points ${}^{abs}P_{w}$.

\paragraph{Referencing plate measurement} As a preparation, some points of the referencing plate have to be measured once. For this purpose, a stereo vision setup (Fig.\@~\ref{fig:stereomeasurement}) is used to determine 3D points. First, the actual positions of the calibration target circles have to be determined. Additionally, there are three reflector nests integrated into the plate, called $n_r, n_g$, and $n_b$ (using color codes in practice). In the stereo vision setup, the ring-shaped surface of the steel reflector nests is matched and their center ${}^{ref}P_{nst,i}, i\in\{n_r,n_g,n_b\}$ is determined relative to the PCS. Assuming the plane that is defined by the reflector nests is parallel to the plate, it is just a z-shift by a known offset $\delta_{nst}$ to retrieve the effective smr position, when mounted:
\begin{align}
    {}^{ref}P_{smr,i} = {}^{ref}P_{nst,i} - 
    \begin{pmatrix}
        0\\0\\\delta_{smr,nst}
    \end{pmatrix}.
\end{align}
When $\delta_{nst}$ is determined as a positive distance, it must be subtracted, because the PCS z-axis mounts into the referencing plate.

\paragraph{Plate pose estimation} For each integrated reflector nest in the referencing plate, i.e., $i\in\{n_r,n_g,n_b\}$, the laser tracker measures the position ${}^{abs}P_{i} := {}^{abs}P_{smr,i}$ of a mounted smr in ACS. The same position is known in the PCS:  ${}^{ref}P_{i} := {}^{ref}P_{smr,i}$. Using the transformation
\begin{align}
    {}^{scn}H_{ref} = {}^{scn}H_{cam} {}^{cam}H_{ref},
\end{align}
it can be transformed into the SCS by
\begin{align}
    {}^{scn}\bar{P}_{i} = {}^{scn}H_{ref} {}^{ref}\bar{P}_{i}.
\end{align}
Matching these points together yields the transformation from SCS to the ACS:
\begin{align}
    {}^{abs}H_{scn} = \underset{{}^{abs}\tilde{H}_{scn}}{\arg\min} \sum_{i\in\{n_r,n_g,n_b\}} \lVert {}^{abs}\bar{P}_{i} - {}^{abs}\tilde{H}_{scn} {}^{scn}\bar{P}_{i}\rVert^2
\end{align}

\paragraph{Robot pose estimation} If the robot is placed on the referencing plate arbitrarily in the world, the measured smr position yields ${}^{abs}P_{rob,0}$ directly, but the basis vectors of the robot orientation ${}^{abs}R_{rob}$ as stated above have to be retrieved. The upward directed normal of the referencing plate surface is
\begin{align}
    {}^{abs}\tilde{\mathbf{n}} &= \left({}^{abs}P_b - {}^{abs}P_r\right) \times \left( {}^{abs}P_g - {}^{abs}P_r \right), \label{eq:robot_pose_nTilde} \\
    {}^{abs}\mathbf{n} &= \frac{{}^{abs}\tilde{\mathbf{n}}}{\lVert{}^{abs}\tilde{\mathbf{n}}\rVert}.
\end{align}
The normal depends on the actual configuration of $n_r, n_g$, and $n_b$. To retrieve the robots heading direction, it must move to a second position. The calculated pose relates to ${}^{abs}P_{rob,0}$, which also relates to the image, which yields ${}^{cam}H_{ref}$. The second robot position ${}^{abs}P_{rob,1}$ is only used to determine the movement vector
\begin{align}
    {}^{abs}\tilde{\mathbf{v}} &= {}^{abs}P_{rob,1} - {}^{abs}P_{rob,0},\\
    {}^{abs}\tilde{\mathbf{v}}_\perp &= {}^{abs}\tilde{\mathbf{v}} - \left( {}^{abs}\tilde{\mathbf{v}} \cdot {}^{abs}\mathbf{n} \right) {}^{abs}\mathbf{n},\\    
    {}^{abs}\mathbf{v}_\perp &= \frac{{}^{abs}\tilde{\mathbf{v}}_\perp}{\lVert{}^{abs}\tilde{\mathbf{v}}_\perp\rVert}.
\end{align}
The orthonormal basis (see Eq.\@~(\ref{eq:rob_R_basisDecomposition})) yields the robot orientation
\begin{align}
    {}^{abs}R_{rob} = 
    \begin{bmatrix} \mathbf{v}_\perp & \mathbf{c} & \mathbf{n} \end{bmatrix} = \begin{bmatrix} {}^{abs}\mathbf{v}_\perp & {}^{abs}\mathbf{c} & {}^{abs}\mathbf{n} \end{bmatrix}
\end{align}
and consequently the complete transformation (see Eq.\@~(\ref{eq:composite_rob_H})):
\begin{align}
    {}^{abs}H_{rob} &=
    \begin{bmatrix}
    {}^{abs}R_{rob} & {}^{abs}P_{rob,0}\\
    0 & 1
    \end{bmatrix},\\
    {}^{rob}H_{abs} &\overset{SE(3)}{=} {}^{abs}H^{-1}_{rob}
\end{align}
Finally, the composite transformation from SCS to RCS is
\begin{align}
    {}^{rob}H_{scn} = {}^{rob}H_{abs} {}^{abs}H_{scn}
\end{align}
and the more intuitive transformation from CCS to RCS is
\begin{align}
    {}^{rob}H_{cam} ={}^{rob}H_{scn} {}^{scn}H_{cam}. \label{eq:rob_H_cam}\enspace .
\end{align}

\section{Mark measurement and experimental setup}
The verification of our method in practice is inspired by our actual use case. The idea is to measure one point on the floor from different perspectives and assume the results to be equal.

In the experiments we use the mobile robot Rita (Fig.~\ref{fig:rita}) equipped with the formerly mentioned $12.2\Unit{MP}$ Camera (see \ref{par:camera}), $3.5\Unit{mm}$ lens mounted. An absolute laser tracker(see \ref{par:lasertracker}) is used for robot positioning (not covered in this work) as well as the measurements of the proposed method and experiments.
\paragraph{Referencing}
A set of camera intrinsic model parameters is expected to be present as stated in Section~\ref{sec:hardware_context}. The referencing method is then executed using the laser tracker and the mobile robot. To minimize side effects from the robot or the environment, the instrument reversal technique is applied analogously to \cite{EVANS1996617}.

\paragraph{Point measurement} The point on the floor is marked with a specific shape (mrk). In the image, the center point $\begin{pmatrix}\mathrm{row_{mrk}},&\mathrm{column_{mrk}}\end{pmatrix}^\top$ of this mark is measured subpixel-precisely by using shape-based matching \citep{Steger2018Book}. Using the rectification map (Eq.\@~(\ref{eq:rectification_map})), it can directly be mapped into the SCS and transformed into the RCS:
\begin{align}
    {}^{rob}\bar{P}_{mrk} = 
    {}^{rob}H_{scn}
    \begin{pmatrix}
        \mathrm{map}^{scn\text{-}xy}_{image}
        \begin{pmatrix}
            \mathrm{row_{mrk}}\\\mathrm{column_{mrk}}
        \end{pmatrix}\\
        0\\1
    \end{pmatrix}.
\end{align}
When the robot pose ${}^{abs}H_{rob}$ is known while taking the measurement image, the abs point is obtained by
\begin{align}
    {}^{abs}\bar{P}_{mrk} = {}^{abs}H_{rob} {}^{rob}\bar{P}_{mrk}.
\end{align}

\paragraph{Floor inclination} As long as no high quality IMU is integrated, the measurement result may be shifted by some floor inclination. Since this shift is related to the floor which is fixed in the ACS, it is independent of the robot pose. Therefore this effect will not be covered by the proposed experiment.

\paragraph{Experiment setup} While $z$, roll, and pitch passively depend on the environment, the $x$ and $y$ position of the robot are actively varied, so that $row_{mrk}$ and $column_{mrk}$ appear at different locations while remaining visible in the image. Employing eight equidistant yaw angles enables a thorough assessment and visualization of calibration errors.

\paragraph{Referencing plate design}
Fig.\@~\ref{fig:referencingplates} shows the referencing plates used in the experiments. 
The assumptions about this plate led to design improvements over time.
In an early stage, there was an in-house-designed wooden referencing plate demonstrator available, see Fig.\@~\ref{fig:woodplate}. 
Large calibration errors could be related to the material and planarity of the calibration target, where the reflector nests were set flush. 
In a further step, these error sources were addressed by designing a glass plate that easily supports the robot weight, see Fig.\@~\ref{fig:glassplate}. 
The reflector nests are placed outside the calibration target so this is not deformed.

\begin{figure}[t]
  \centering
  \subfigure[\label{fig:p0alt}]{
    \includegraphics[width=0.45\linewidth]{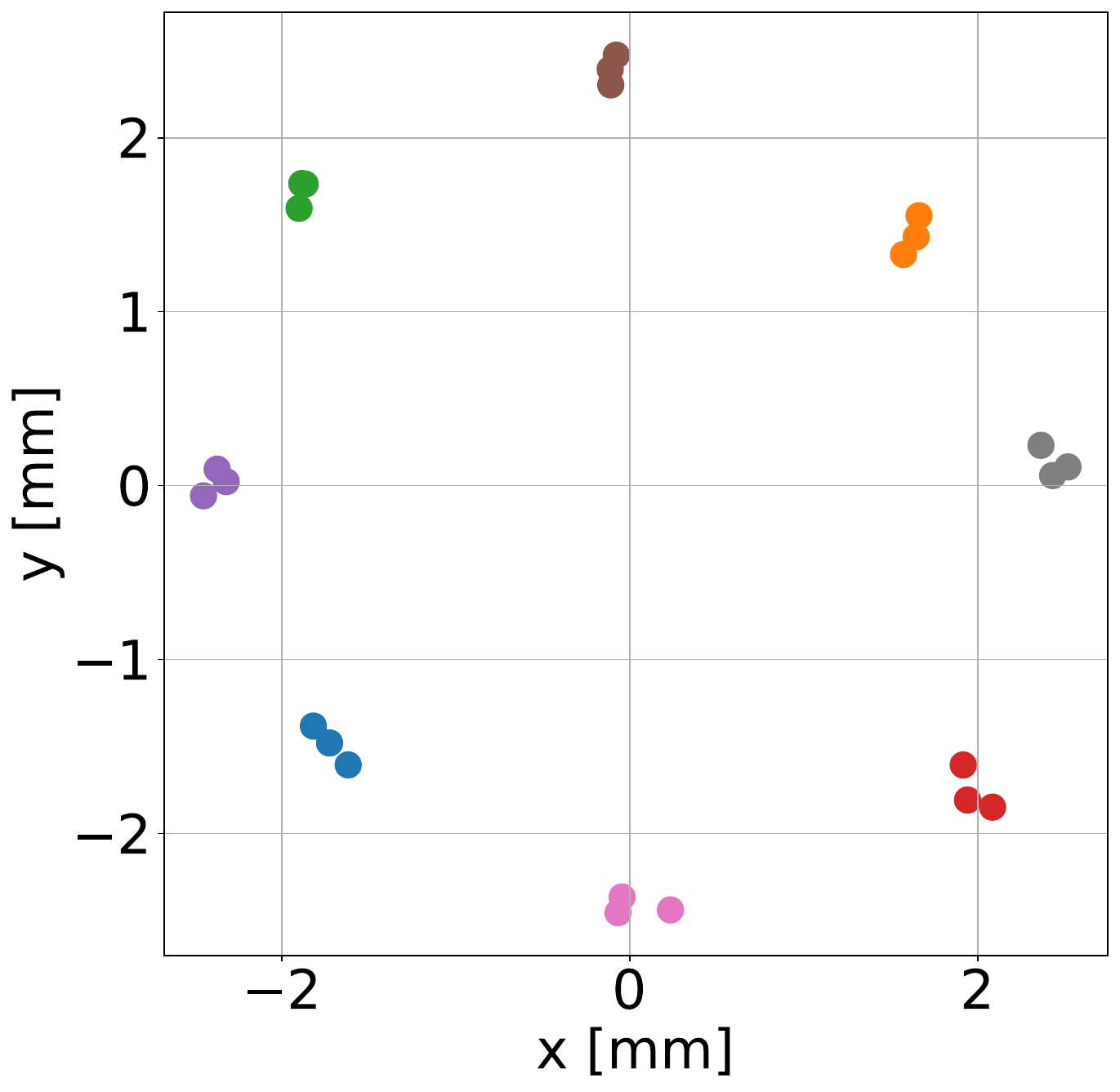}
  }
  \subfigure[\label{fig:p1alt}]{
    \includegraphics[width=0.45\linewidth]{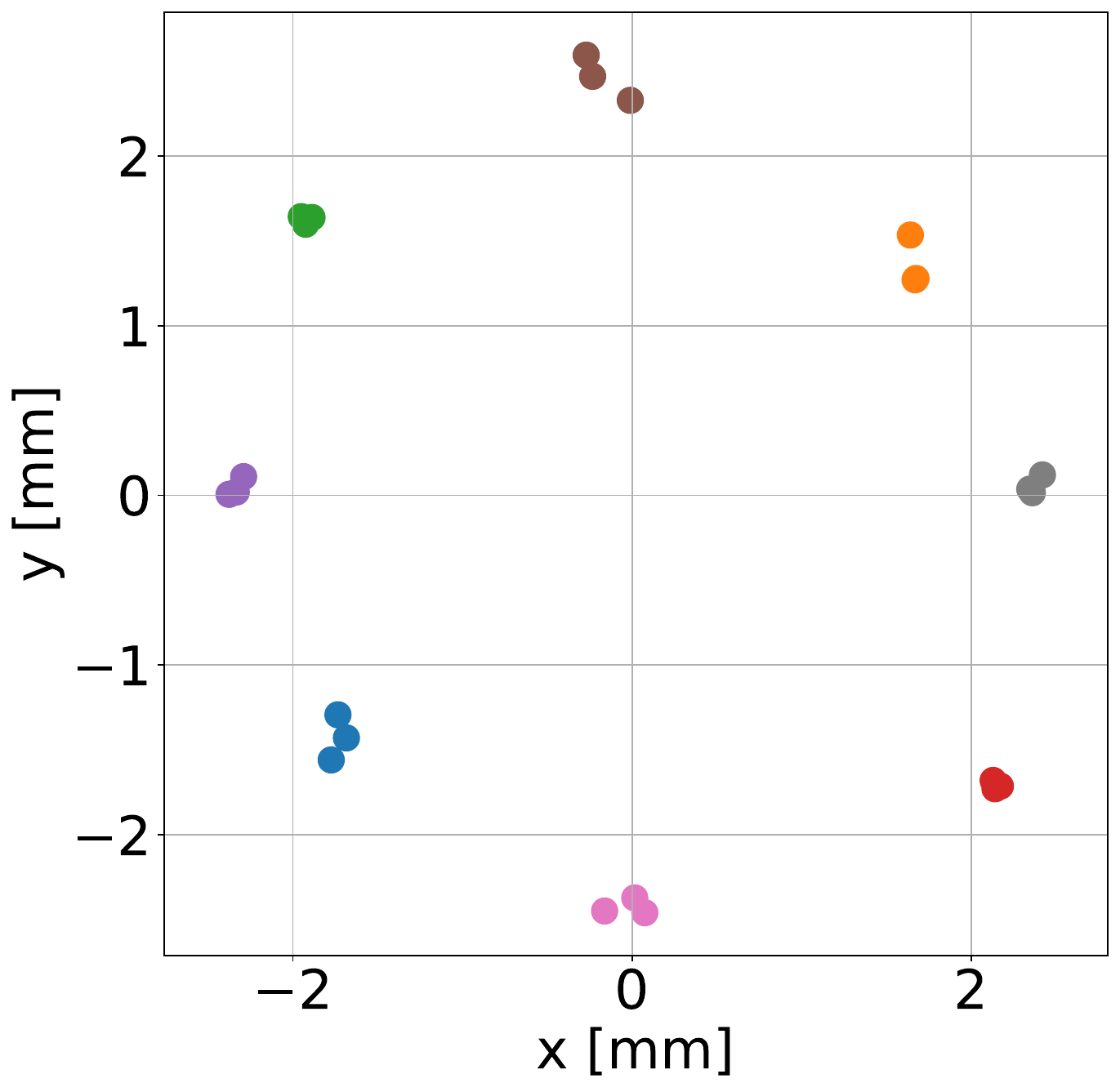}
  }
  \subfigure[\label{fig:p0neu}]{
    \includegraphics[width=0.45\linewidth]{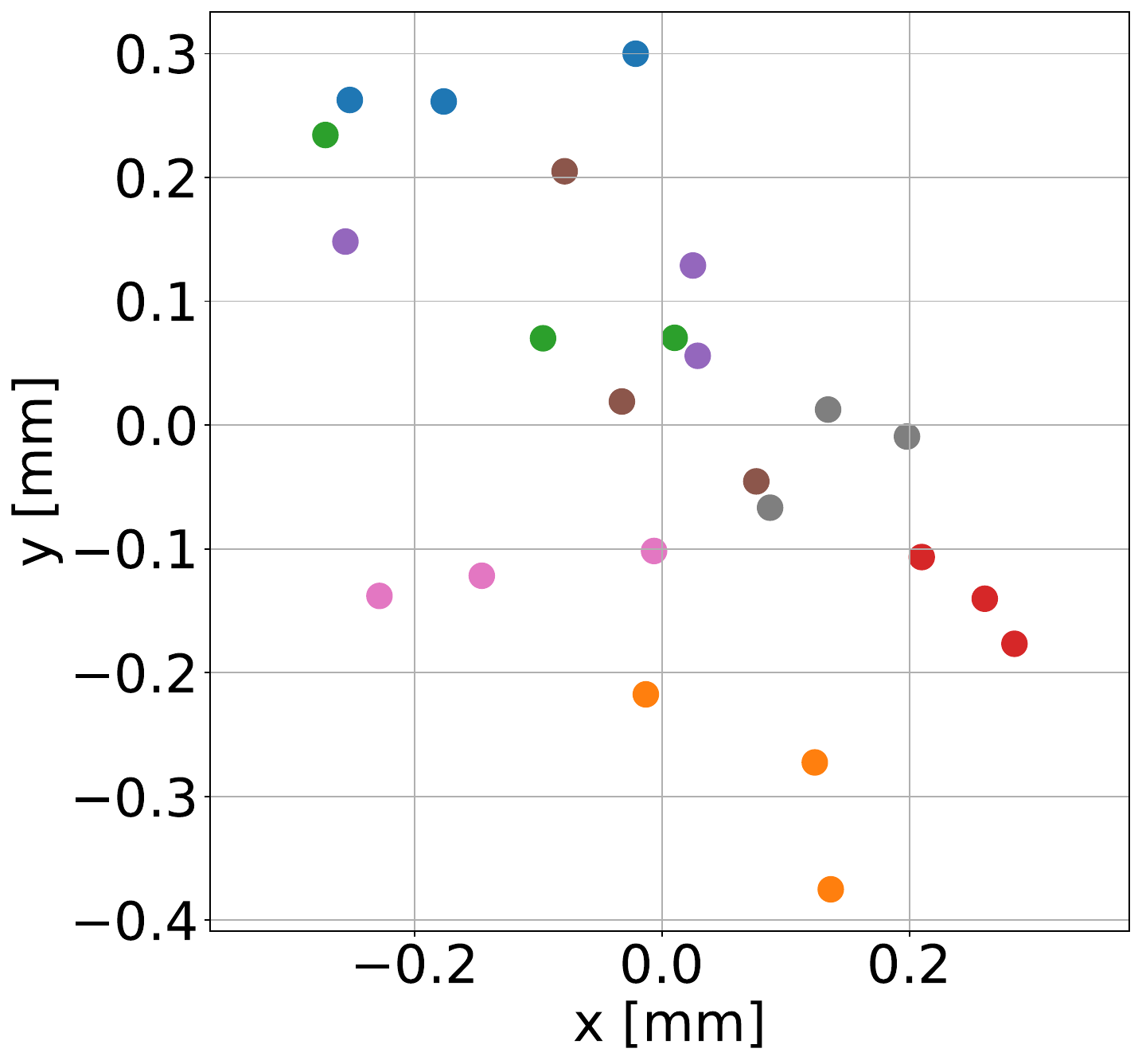}
  }
  \subfigure[\label{fig:p1neu}]{
    \includegraphics[width=0.45\linewidth]{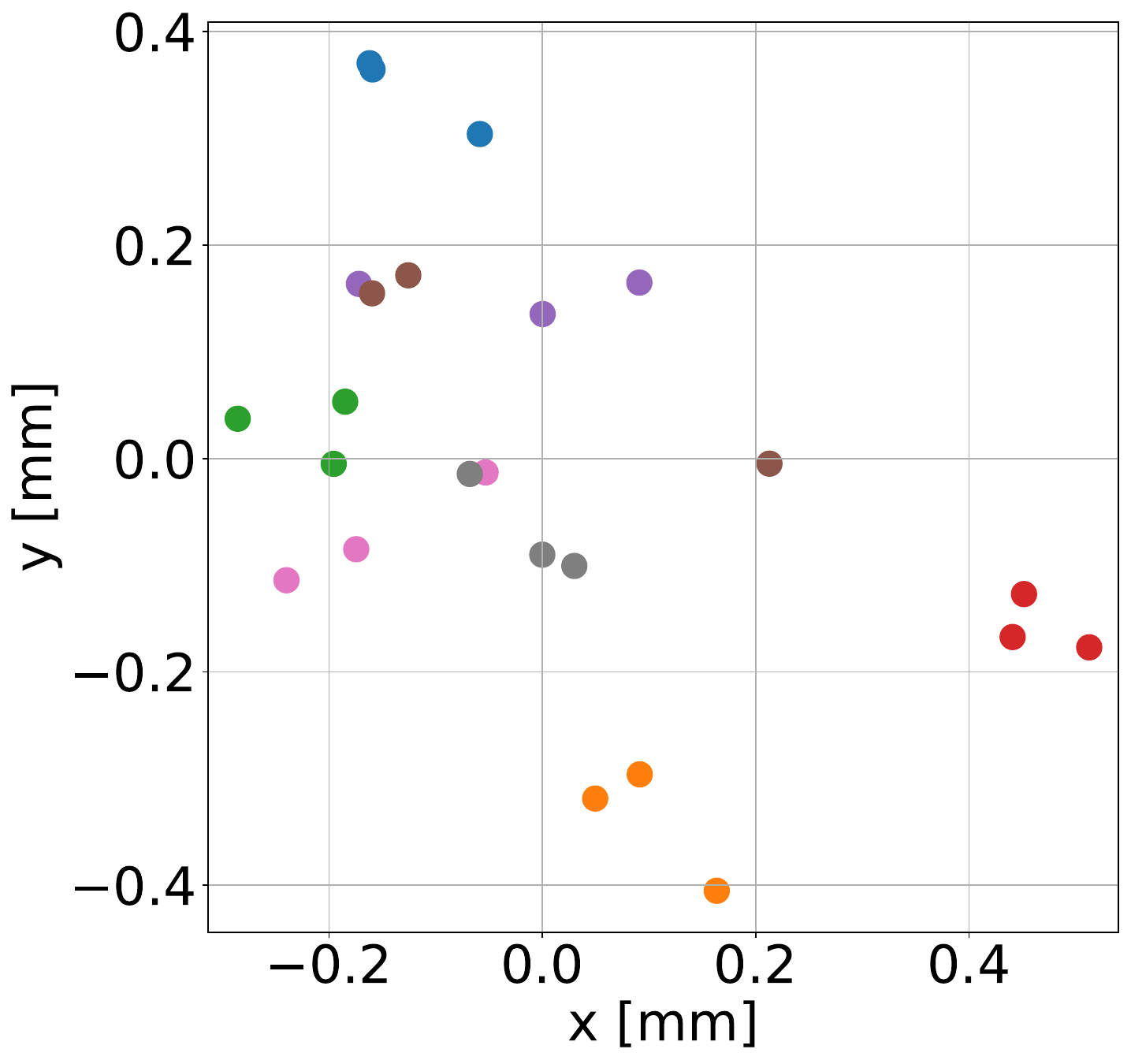}
  }
  \subfigure{
    \includegraphics[width=0.7\linewidth]{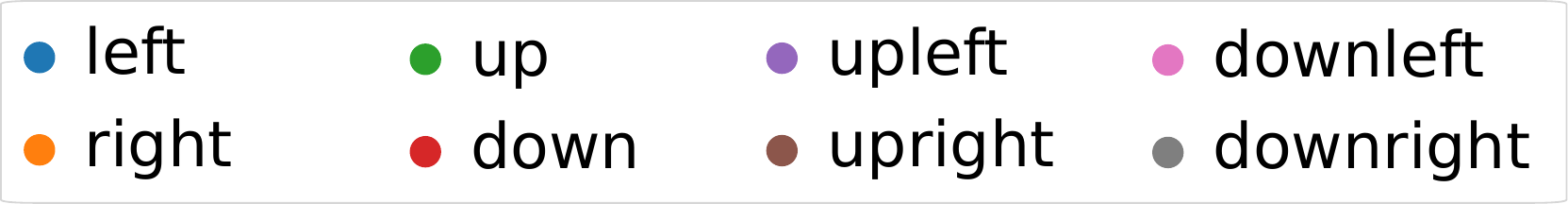}
  }
  \caption{Results of repeated absolute measurements of marks on the floor in an experiment. Each plot is normalized to the overall mean position, see Table~\ref{tab:target_direction_metrics}. (a) and (c) denote the same point as Table~\ref{tab:target_direction_metrics} does. (b) and (d) denote another mark on the floor. (a) and (b) are based on the in-house-designed wooden referencing plate. (c) and (d) are based on the bespoke glass referencing plate}
  \label{fig:results}
\end{figure}

\begin{table*}[bt]
\centering
\begin{tabular}{lcccccc}
\hline
\textbf{Direction} & \multicolumn{2}{c}{\textbf{Mean Position [mm]}} & \multicolumn{3}{c}{\textbf{Distance Metrics [mm]}} & \textbf{Approach Angle Range [$^\circ$]} \\
 & X & Y & Max from Mean & Mean from Mean & Cluster Radius & \\
\hline
left       & 130825.822 & 8525.577 & 0.131 & 0.098 & 0.117 & [89.48, 90.66] \\
right      & 130826.054 & 8525.014 & 0.119 & 0.094 & 0.109 & [-90.40, -89.96] \\
up         & 130825.853 & 8525.427 & 0.188 & 0.140 & 0.163 & [0.12, 0.36] \\
down       & 130826.224 & 8525.161 & 0.054 & 0.042 & 0.052 & [179.53, 179.68] \\
upleft     & 130825.905 & 8525.413 & 0.192 & 0.139 & 0.150 & [45.00, 45.06] \\
upright    & 130825.960 & 8525.362 & 0.160 & 0.124 & 0.148 & [-45.11, -44.87] \\
downleft   & 130825.845 & 8525.182 & 0.122 & 0.093 & 0.113 & [135.27, 135.43] \\
downright  & 130826.112 & 8525.281 & 0.070 & 0.056 & 0.063 & [-135.27, -134.47] \\
\hline
\textbf{All} & 130825.972 & 8525.302 & 0.399 & 0.239 & 0.374 &  \\
\hline
\multicolumn{7}{l}{\textit{Mean L2-distance between cluster means}: 0.215 mm}
\end{tabular}

\caption{Quantitative results of repeated absolute measurements of marks
on the floor in an experiment. Individual (non-averaged) measured positions of this experiment are plotted in \ref{fig:p0neu}. \emph{Mean position} is the mean of the measured absolute coordinates of the respective cluster. \emph{Max. and mean from mean} denote the (max. and mean, respectively) distance to this \emph{mean position}. \emph{Cluster diameter} is the max. distance between two points in the cluster. In the last row, those metrics are calculated for \emph{all} directions together.}
\label{tab:target_direction_metrics}
\end{table*}

\section{Results}\label{sec:results}
\paragraph{Wooden demonstrator referencing plate}
Using the wooden referencing plate, which especially breaks assumptions about the planarity, our experiments show a major calibration error. Repeated measurements of one printed mark on the floor show a systematic error depending on the robot heading. The robot approached the mark from eight different directions, an image was taken, and the position of the mark was measured and calculated consequently. The measurement results are clustered in eight regions, depending on the approach direction as shown in Figs.\@~\ref{fig:p0alt} and \ref{fig:p1alt} for two different printed marks on the floor. 

\paragraph{Bespoke glass referencing plate}
Figs.\@~\ref{fig:p0neu} and \ref{fig:p1neu} show the same experiment for the glass plate. The same clusters are also distinguishable, but Tab.\@~\ref{tab:target_direction_metrics} shows the diameter, calculated from the radius $d_i = 2 r_i$ with $i$ varying over the eight directions, of the enclosing circle of all measurements $d_{All} = 0.747\Unit{mm}$ is significantly below a millimeter, while the cluster diameters 
%themselves 
range from $d_{Down} = 0.103\Unit{mm}$  to $d_{Up} = 0.326\Unit{mm}$ .

\section{Discussion}\label{sec:discussion}
\paragraph{}
For the in-house-designed referencing plate, Figures\@~\ref{fig:p0alt} and \ref{fig:p1alt} show the expected disposition of the clusters for a wrong transformation between camera and robot. The measurements are equally shifted away from each direction. If a systematic error in the transformation between camera and robot ${}^{rob}H_{cam}$, see Eq.\@~(\ref{eq:rob_H_cam}), is dominant, this circular arrangement of clusters is expected. In this case, there are some assumptions missed, which can cause an error in ${}^{rob}H_{cam}$. Since the glass plate successfully addresses these assertions as stated above, these clusters do not show this systematic error. The calibration has reached a sub-mm level and does not remain the dominant error source in the proposed experiment.

\paragraph{}
The most obvious error source is about the missing robot pitch and roll during the test measurements. Taking the normal vector of the referencing plate to determine the robot's pitch and roll (see Eq.\@~(\ref{eq:robot_pose_nTilde})) there is no need for an intrinsic measurement of the robot rotation during the referencing method. The proposed experiment to verify the referencing method is comparing altered, repeated measurement results against each other, where no absolute ground truth is involved. However, when a mark on the floor is measured, the robot's roll and pitch have also to be taken into account to obtain the real absolute coordinate of this mark. Unfortunately, no IMU is yet available which delivers high quality rotation information. Applying an IMU will therefore not change the referencing method itself, but the test, in which the absolute position of a mark is checked in the ACS, will be affected. Then the method can be verified by comparing an absolute measurement against an absolute ground truth.

\section{Conclusion}\label{sec:conclusion}
When a mobile robot possesses a ground-facing camera whose orientation must be known in the robot frame to make high precision measurements, even state-of-the-art methods like hand--eye calibration show major disadvantages. In this work, we proposed an alternative solution and showed that the proposed methodology is feasible.
A special referencing plate was engineered to connect the optically defined PCS via three reflector nests to the world-defined ACS. Using this referencing plate, the absolute camera pose as well as the absolute robot pose can be retrieved. Combining these poses yields the camera pose w.r.t.\ the robot.
The results have shown errors in sub-mm range which will be further improved in future by involving an IMU.

\section*{Acknowledgments}
This Project is supported by the Federal Ministry for Economic Affairs and Climate Action (BMWK) on the basis of a decision by the German Bundestag.

\newpage

{
	\begin{spacing}{1.17}
		\normalsize
		\bibliography{bibliography} % Include your own bibliography (*.bib), style is given in isprs.cls
	\end{spacing}
}

\end{document}